# SpecXNet: A Dual-Domain Convolutional Network for Robust Deepfake Detection

Inzamamul Alam
Department of Computer Science and Engineering
Sungkyunkwan University
Suwon, Republic of Korea
inzi15@g.skku.edu

Md Tanvir Islam
Department of Computer Science and Engineering
Sungkyunkwan University
Suwon, Republic of Korea
tanvirnwu@g.skku.edu

Simon S. Woo*
Department of Artificial Intelligence
Sungkyunkwan University
Suwon, Republic of Korea
swoo@g.skku.edu

## Abstract

The increasing realism of content generated by GANs and diffusion models has made deepfake detection significantly more challenging. Existing approaches often focus solely on spatial or frequency-domain features, limiting their generalization to unseen manipulations. We propose the Spectral Cross-Attentional Network (SpecXNet), a dual-domain architecture for robust deepfake detection. The core **Dual-Domain Feature Coupler (DDFC)** decomposes features into a local spatial branch for capturing texture-level anomalies and a global spectral branch that employs Fast Fourier Transform to model periodic inconsistencies. This dual-domain formulation allows SpecXNet to jointly exploit localized detail and global structural coherence, which are critical for distinguishing authentic from manipulated images. We also introduce the **Dual Fourier Attention (DFA)** module, which dynamically fuses spatial and spectral features in a content-aware manner. Built atop a modified XceptionNet backbone, we embed the DDFC and DFA modules within a separable convolution block. Extensive experiments on multiple deepfake benchmarks show that SpecXNet achieves state-of-the-art accuracy, particularly under cross-dataset and unseen manipulation scenarios, while maintaining real-time feasibility. Our results highlight the effectiveness of unified spatial-spectral learning for robust and generalizable deepfake detection. To ensure reproducibility, we released the full code on GitHub.

## CCS Concepts

• **Computing methodologies** → **Computer vision tasks**.

## Keywords

Deepfake Detection, Dual-Domain Learning, Fake Image Classification, Frequency Domain, Security

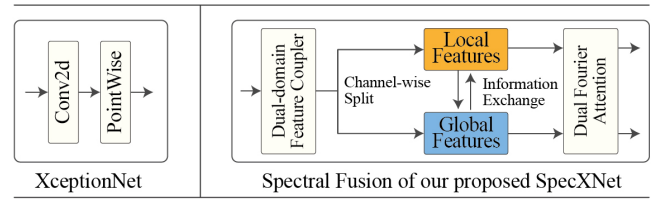

**Figure 1: High-level architectural comparison between the original XceptionNet and our proposed modification. While XceptionNet utilizes standard depthwise separable convolutions (Conv2d + PointWise), our enhanced version decomposes each block into dual pathways: a local spatial branch using conventional convolutions and a global spectral branch powered by frequency-aware operations. These branches are then adaptively fused using the proposed Dual Fourier Attention (DFA) module to jointly exploit fine-grained local details and holistic spectral cues.**

## 1 Introduction

The proliferation of generative AI techniques such as Generative Adversarial Networks (GANs) [34] and Diffusion [57, 60] models has made the creation of highly realistic synthetic images both accessible and efficient. While these technologies offer creative potential, they also introduce significant societal risks, including misinformation, identity fraud, and political manipulation [49]. This has led to an urgent demand for robust and generalizable deepfake detection systems capable of operating across diverse image domains and generative models.

A range of deep learning methodologies has been proposed [5, 28, 40, 50, 53, 61, 62, 66, 68, 69, 69, 72] to address the deepfake detection problem, including preprocessing techniques designed to extract power spectral features for classification [16, 67]. Despite their initial success, these approaches often exhibit reduced effectiveness when applied to content generated by more recent and sophisticated generative models [2, 21, 56, 75]. In parallel, various forensic analysis techniques [17] have been explored; however, many existing fake image detection frameworks still lack the architectural robustness and accuracy necessary to reliably identify the latest AI-generated forgeries.

Recent state-of-the-art (SOTA) methods [3, 11, 15, 51, 64, 65] have explored various avenues to address this challenge. Deepfake detection is commonly approached as a supervised classification task, where neural networks are trained to distinguish between genuine and altered visual content [11, 29, 47, 51]. However, a major limitation lies in their generalization capability of the models trained solely on specific types of synthetic images often struggle to accurately detect previously unseen manipulation techniques [29, 37]. Vision-language models (VLMs) like CLIP [55] have demonstrated impressive zero-shot capabilities for fake image detection by leveraging large-scale visual and textual representations. Adaptation strategies such as prompt tuning and adapter networks have enhanced CLIP's performance and generalizability across GAN and diffusion-based datasets [39, 73]. However, these approaches remain dependent on large-scale training and inference resources, often overlooking the energy efficiency and temporal dynamics crucial for real-time or edge deployment scenarios.

Simultaneously, frequency-based methods like UGAD [4] have attempted to harness unique spectral signatures of AI-generated images through Radial Integral Operations and Spatial Fourier Extraction. These techniques effectively distinguish real from fake images using frequency-domain fingerprints, achieving competitive accuracy across a variety of generative methods. However, such systems also rely on deep convolutional backbones like ResNet152 [25], which can be computationally expensive and may not fully exploit temporal information latent in the input data stream. Moreover, most existing methods operate in either the spatial or frequency domain, treating the two sources of evidence independently. This separation often causes subtle manipulation traces that are only evident in one domain to be overlooked.

To address these issues, we present SpecXNet, a deepfake detection framework that enhances the XceptionNet backbone by integrating spatial and spectral representations within a unified convolutional architecture. Unlike the original XceptionNet, SpecXNet introduces a dual-branch design: a local spatial branch that captures fine-grained textures via standard convolutions, and a global spectral branch that applies two-dimensional Fast Fourier Transform to model long-range frequency patterns. These complementary branches are fused through a novel Dual Fourier Attention (DFA) mechanism, which generates channel-wise attention maps to enable reciprocal modulation and content-adaptive fusion. This design promotes synergistic learning of spatial textures and spectral anomalies, significantly improving the model's ability to detect subtle traces of manipulation. SpecXNet embeds this dual-domain architecture within modified depthwise separable convolution blocks, referred to as Dual-Domain Feature Couplers (DDFC). As illustrated in Figure 1, these enhanced blocks offer a lightweight and scalable solution while delivering superior expressive power compared to the vanilla Xception architecture.

- Firstly, we propose a Dual-Domain Feature Coupler (DDFC) deepfake detection framework that jointly models spatial detail and spectral context through a bifurcated convolutional architecture. This enables the learning of robust multi-scale representations for better manipulation detection.
- Secondly, we introduce a novel Dual Fourier Attention (DFA) module that adaptively fuses local and global features using cross-domain modulation and attention-weighted integration, enhancing the semantic alignment of spatial and frequency cues.
- We demonstrate the effectiveness and generalization ability of our proposed SpecXNet through extensive evaluations of diverse generative models and benchmark datasets, showing consistent performance improvements over existing spatial and spectrum-based approaches.

## 2 Related Works

A wide range of approaches have been proposed in recent years to address the challenge of deepfake detection [5, 6, 12–14, 24, 28–31, 36, 38, 40, 41, 53, 61, 66, 68, 69]. Traditional supervised learning methods have been widely adopted, where models are trained to differentiate between authentic and manipulated content. For instance, Wu et al. [65] adopted a conventional classification strategy, while Cozzolino et al. [18] introduced spectral preprocessing techniques that extract power spectrum features to enhance classification performance. Language-guided approaches have also emerged, with Radford et al. [55] leveraging perceptual learning guided by text supervision, and Wang et al. [65] presenting the DIRE model designed explicitly for diffusion-generated image detection. Similarly, Cozzolino et al. [18] proposed a GAN detection method based on a ResNet152 backbone. In another line of work, Zhang et al. [70] utilized Discrete Cosine Transform (DCT) to analyze power spectrum properties, though their evaluation was limited to a single diffusion model. Jeong et al. [32] employed spectral analysis by training generator-discriminator pairs and analyzing the resultant power spectrum distribution. In parallel, Radial Integral Operation (RIO) has been used to accumulate power spectrum density across radii, with findings indicating that deepfake images tend to exhibit consistent spectral distributions. This observation can be exploited for classification. Beyond spectral techniques, other forensic strategies have been explored. Corvi et al. [17] investigated noise print-based camera fingerprinting, while Mandelli et al. [46] focused on distinguishing real and manipulated Western blot images, albeit with limited generality. Ma et al. [43] applied a combination of statistical analysis and neural networks to detect synthesis artifacts in generative models. Unlike previous methods, our model unifies spatial and spectral analysis through a dual-branch architecture with adaptive attention, enabling robust and generalizable deepfake detection across both GAN- and diffusion-based manipulations.

## 3 Proposed Method: SpecXNet

We introduce the Spectral Cross-Attentional Network (SpecXNet), a convolutional neural network (CNN) architecture that jointly models local spatial and global spectral representations. As illustrated by the cyan block in Figure 2, the core design of SpecXNet lies in the explicit partitioning of input features into two complementary branches: one operating in the spatial domain and the other in the frequency domain. This separation enables the network to simultaneously capture fine-grained local patterns and broad contextual cues by applying standard convolutions to spatial features and Fast Fourier Transform to spectral features. The dual-domain architecture significantly improves the model's capacity to detect diverse manipulation artifacts across multiple scales and content types.

### 3.1 Dual-Domain Feature Coupler (DDFC)

Given an input feature tensor $X \in \mathbb{R}^{C \times H \times W}$, our proposed DDFC initiates a systematic decomposition of $X$ to effectively leverage distinct yet complementary representations. As depicted by the violet block in Figure 2, we strategically partition the input into two specialized subsets: local spatial features $X_l$ and global spectral features $X_g$. Formally, this decomposition is defined as follows:

$$(X_l, X_g) = \mathcal{D}(X; \alpha), \quad (1)$$

where $\mathcal{D}(\cdot)$ signifies the channel-wise splitting operator governed by the hyperparameter $\alpha \in [0, 1]$, carefully selecting the proportion of channels designated for spectral analysis. Thus, $X_l \in \mathbb{R}^{(1-\alpha)C \times H \times W}$ is tailored towards capturing localized spatial intricacies, while $X_g \in \mathbb{R}^{\alpha C \times H \times W}$ is preserved explicitly for global-scale spectral domain processing.

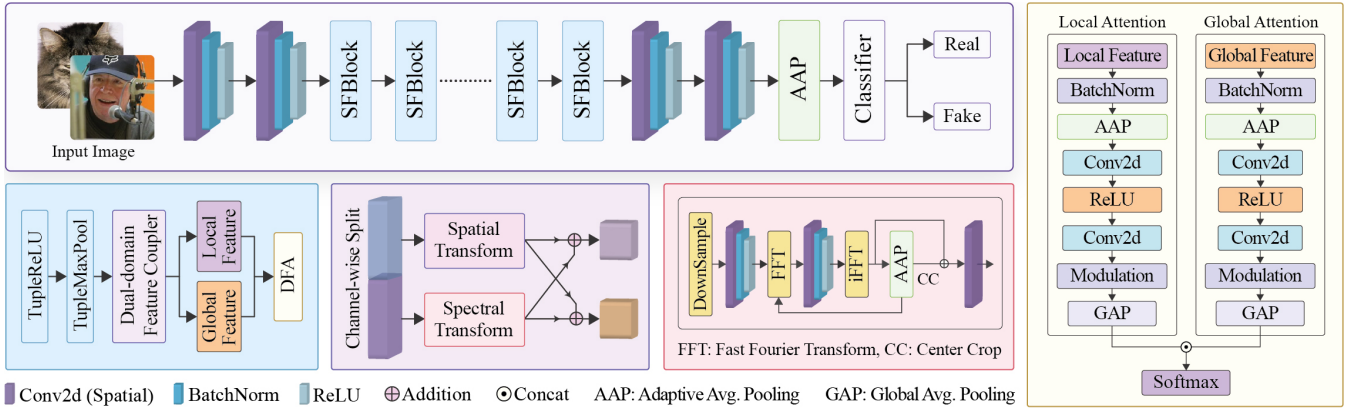

Figure 2: Overview of the proposed SpecXNet architecture for deepfake detection. The top panel presents the full pipeline, where an input image is processed through a sequence of Spectral Fusion Blocks (SFBlocks) and adaptive average pooling before final classification. The bottom-left panel illustrates the internal structure of each SFBlock, which consists of spatial convolution, max-pooling, and a Dual-Domain Feature Coupler (DDFC) module that splits the feature map into local and global branches using spatial and spectral domains, respectively. These are modulated and fused using the Dual Fourier Attention (DFA) module. The bottom-middle panel details the internal structure of the DDFC, showing how spatial and spectral pathways are handled via spatial convolution and Fourier-based spectral transforms. The bottom-right panel visualizes the internal design of this Spectral Transform, which performs downsampling, applies Fast Fourier Transform (FFT) and inverse FFT to process frequency-domain information, and incorporates a low-frequency residual stream through adaptive average pooling (AAP) and center cropping (CC) to enhance frequency-selective representation. Lastly, the right panel provides a breakdown of the DFA module, which computes cross-domain attentions from both branches, applies residual modulation, and combines them using a learned softmax-weighted fusion for final representation.

**Local Spatial Branch.** The core aim of the local spatial branch is to meticulously preserve and capture the intricate spatial patterns, subtle edges, and fine textures embedded within the input data. To this end, we utilize CNN layers, renowned for their efficacy in extracting local visual patterns. Given convolutional kernels parameterized by weights $W_l$ and biases $b_l$, we perform the convolution operation on the local features $X_l$. The precise mathematical expression is formulated as follows:

$$Z_l = W_l * X_l + b_l = \sum_i \sum_j W_l(i, j) X_l(h-i, w-j) + b_l. \quad (2)$$

Following the convolutional transformation, we implement a batch normalization (BN) procedure to stabilize the intermediate activations and accelerate convergence as follows:

$$\text{BN}(Z_l) = \gamma \frac{(Z_l - \mu_B)}{\sqrt{\sigma_B^2 + \epsilon}} + \beta, \quad (3)$$

where $\mu_B$ and $\sigma_B$ respectively represent the mean and variance computed across mini-batches, $\gamma$ and $\beta$ denote learnable scaling and shifting parameters, and $\epsilon$ is a small numerical stability constant. Lastly, we incorporate a non-linear rectified linear unit (ReLU) activation $\sigma(\cdot)$ to introduce nonlinearity into our representation:

$$Y_l = \sigma\left(\text{BN}(Z_l)\right) = \max\left(0, \text{BN}(Z_l)\right). \quad (4)$$

The resultant feature map $Y_l$ explicitly encodes local structures, laying a robust foundation for capturing high-resolution spatial details crucial to accurate representation learning.

**Global Spectral Branch.** Simultaneously, to integrate a broader receptive field and global contextual understanding, we propose an innovative spectral processing branch. This global branch explicitly leverages frequency-domain representation via the Fast Fourier Transform (FFT), fundamentally transforming spatial data into frequency-domain representations. We initiate the process by converting spatial domain features $X_g$ into frequency domain representations $X_g^{\mathcal{F}}$ via a 2D Fourier Transform, mathematically expressed as follows:

$$X_g^{\mathcal{F}}(u, v) = \sum_{h=0}^{H-1} \sum_{w=0}^{W-1} X_g(h, w)\, e^{-j2\pi\left(\frac{uh}{H} + \frac{vw}{W}\right)}, \quad (5)$$

where $(u, v)$ indicates the frequency-domain indices, and the transform effectively encodes global-level correlations across the entirety of the input spatial field into each spectral coefficient.

After obtaining spectral representations from Eq. 5, we apply a learnable modulation operation $\Phi(\cdot)$, which serves to selectively emphasize critical spectral features. This operation is implemented via an element-wise spectral filtering parameterized by learnable spectral weights $W_g$. Hence, the modulated spectral coefficients $X'_g F(u, v)$ are represented as follows:

$$X'_g F(u,v) = \sigma\left(\text{BN}(W_g \odot X_g^{\mathcal{F}}(u,v))\right), \quad (6)$$

where $\odot$ denotes the Hadamard (element-wise) product, and the spectral batch normalization is employed analogously to its spatial counterpart to facilitate stable learning dynamics.

To complete the spectral processing cycle, we subsequently revert the frequency-modulated features $X'_g F(u, v)$ back into the spatial domain. This inversion is accomplished via the inverse Fourier Transform, meticulously reconstructing a spatial representation $Y_g(h, w)$ from the global-scale spectral information. Explicitly, the inverse Fourier Transform is formulated as:

$$Y_g(h,w) = \frac{1}{HW} \sum_{u=0}^{H-1} \sum_{v=0}^{W-1} X'_g F(u,v)\, e^{j2\pi\left(\frac{uh}{H} + \frac{vw}{W}\right)}. \quad (7)$$

The resulting global representation $Y_g$ thus inherently encodes comprehensive contextual information spanning the entire input, effectively complementing the local branch by capturing extensive inter-pixel dependencies that exceed the limited receptive field achievable through purely local convolutions. As illustrated in **red block** Figure 2, the spectral branch leverages both full-resolution and pooled low-frequency signals via the *Spectral Transform* module, integrating them through frequency-domain processing and subsequent refinement. Collectively, the joint processing of local and global branches delivers a synergistic integration of fine-grained local information and expansive global context, offering a robust representation for complex pattern recognition tasks.

### 3.2 Dual Fourier Attention (DFA)

To effectively leverage the complementary yet distinct representations learned by the local and global processing branches, we introduce a meticulously designed Dual Fourier Attention (DFA) mechanism. DFA is explicitly formulated to adaptively fuse spatially precise local features with contextually enriched global spectral representations. This fusion mechanism dynamically recalibrates feature importance across domains, significantly enhancing the discriminative capabilities of the resultant representations.

The proposed DFA commences with the compression of spatial information into compact feature descriptors via a global average pooling (GAP) operation. Given spatial feature maps $Y_l \in \mathbb{R}^{C_l \times H \times W}$ and $Y_g \in \mathbb{R}^{C_g \times H \times W}$ from the local and global branches respectively, the GAP operation aggregates spatial information into concise global descriptors $Z_l \in \mathbb{R}^{C_l}$ and $Z_g \in \mathbb{R}^{C_g}$ as follows:

$$Z_l = \mathcal{P}(Y_l) = \frac{1}{HW} \sum_{h=1}^{H} \sum_{w=1}^{W} Y_l(:, h, w), \quad (8)$$

$$Z_g = \mathcal{P}(Y_g) = \frac{1}{HW} \sum_{h=1}^{H} \sum_{w=1}^{W} Y_g(:, h, w), \quad (9)$$

where $\mathcal{P}(\cdot)$ denotes the global spatial averaging across all spatial locations for each feature channel independently. Such aggregation succinctly encapsulates dominant spatial patterns and global spectral structures, distilling the crucial representational insights from both local and global domains into lower-dimensional embeddings.

Subsequent to global pooling, these feature descriptors are then utilized to generate two attention maps, specifically tailored for bidirectional cross-domain modulation. The global attention map $A_g \in \mathbb{R}^{C_l}$ and local attention map $A_l \in \mathbb{R}^{C_g}$ are derived from the respective feature descriptors through separate linear transformations with trainable parameters as follows:

$$A_g = \sigma(W_a Z_g + b_a), \quad (10)$$

$$A_l = \sigma(W_b Z_l + b_b), \quad (11)$$

where $W_a \in \mathbb{R}^{C_l \times C_g}$, $W_b \in \mathbb{R}^{C_g \times C_l}$, $b_a \in \mathbb{R}^{C_l}$, and $b_b \in \mathbb{R}^{C_g}$ represent learnable weights and biases respectively, initialized to optimize feature fusion during training. Here, the sigmoid activation function $\sigma(x) = 1/(1 + e^{-x})$ is strategically employed to ensure that the computed attention maps $A_g$ and $A_l$ smoothly range between 0 and 1, providing a probabilistic interpretation of the relative importance of features from each domain.

*(In practice, this attention generation is implemented via two lightweight convolutional sub-networks composed of $1 \times 1$ convolutions with intermediate non-linearity (ReLU), closely mimicking fully connected layers applied channel-wise. These convolutional pathways are applied to the GAP-reduced tensors reshaped as single-spatial pixel feature maps.)*

*(Furthermore, to ensure shape consistency during fusion, the attention maps $A_g$ and $A_l$ are upsampled via bilinear interpolation if their spatial dimensions differ from their target domains. This resizing ensures proper broadcasting during modulation.)*

The cross-domain modulation phase explicitly utilizes the derived attention maps to modulate the complementary domain features through element-wise multiplications. This crucial step selectively emphasizes or suppresses features according to cross-domain feedback, effectively integrating multi-scale information into each domain's representation. The modulated feature maps are thus formulated as follows:

$$\tilde{Y}_l = Y_l \odot A_g, \quad (12)$$

$$\tilde{Y}_g = Y_g \odot A_l, \quad (13)$$

where $\odot$ is element-wise multiplication, which dynamically adjusts the feature intensity based on domain-specific attention scores.

*(Notably, in our implementation, a residual formulation is adopted where the attention output is added back to the original features, i.e., $\tilde{Y}_l = Y_l + Y_l \odot A_g$ and $\tilde{Y}_g = Y_g + Y_g \odot A_l$, to preserve identity information and stabilize gradient flow. This residual modulation enhances representation fidelity during training.)*

Through this adaptive modulation, DFA robustly enables the local branch to explicitly incorporate global context and simultaneously allows the global branch to embed refined local details.

To finalize the fusion, we introduce a carefully designed adaptive weighting strategy that further optimizes the combined contribution of each modulated feature map. To achieve this, we concatenate the pooled global descriptors $Z_l$ and $Z_g$ into a single unified vector and subsequently compute adaptive fusion coefficients using a learnable linear mapping that is defined as follows:

$$[\gamma_l, \gamma_g] = \text{softmax}(W_f [Z_l; Z_g] + b_f), \quad (14)$$

where $[\cdot;\cdot]$ denotes concatenation, $W_f \in \mathbb{R}^{2 \times (C_l + C_g)}$ and $b_f \in \mathbb{R}^2$ are trainable parameters, and the softmax function ensures that the weights $\gamma_l$ and $\gamma_g$ represent valid probability distributions, thereby guaranteeing their sum to unity. The learned coefficients intuitively adapt to input feature characteristics, allowing the network to intelligently balance the contribution from each domain based on the prevailing representational context. Ultimately, the final integrated representation $Y_{\text{out}}$ is computed as a weighted combination of the modulated local and global feature maps using these adaptive fusion weights:

$$Y_{\text{out}} = \gamma_l \tilde{Y}_l + \gamma_g \tilde{Y}_g. \tag{15}$$

As illustrated by the **orange block** in Figure 2, the DFA mechanism captures the interactions between local and global features through symmetric attention pathways, followed by residual modulation and adaptive fusion driven by global descriptors. This meticulous formulation endows DFA with the capacity to dynamically recalibrate and optimally merge local and global representations in a context-sensitive manner. Consequently, the DFA enhances the representational capability of SpecXNet, facilitating superior generalization performance and enriched feature expressiveness.

### 3.3 SpecXNet Backbone

**XceptionNet Integration Procedure.** We integrate our proposed DDFC and DFA mechanisms into a refined XceptionNet architecture. The primary innovation behind XceptionNet lies in the systematic employment of depthwise separable convolutions, which factorize standard convolutional operations into spatially separate depthwise and pointwise convolutions that can be delineated as follows:

$$Y_{sep} = W_p * (W_d \circledast X), \tag{16}$$

where $X$ denotes the input feature map, $W_d$ represents a depthwise convolution kernel applied independently to each channel, $\circledast$ denotes depthwise convolution, and $W_p$ indicates a pointwise convolution kernel responsible for inter-channel interactions.

Building upon this efficient convolutional design, we carefully embed our dual-domain modules: local spatial branch and global spectral branch, interconnected by our proposed DFA, into XceptionNet blocks. Each enhanced Xception block systematically processes the input feature maps $X^i \in \mathbb{R}^{C \times H \times W}$ through a balanced interplay between local and global transformations. Specifically, for the $i$-th modified block, input features are initially decomposed into local $(X_l^i)$ and global $(X_g^i)$ components using the same channel-wise decomposition described in Eq. 1.

The local features $X_l^i$ are processed using depthwise separable convolutions, batch normalization, and nonlinear activation, following a structure similar to Eqs. 2–4. Simultaneously, the global spectral features $X_g^i$ undergo frequency-domain transformation, modulation, and reconstruction. The forward Fourier Transform is computed as shown in Eq. 5, followed by learnable spectral modulation via Eq. 6, and finally, the inverse Fourier Transform is applied as per Eq. 7 to retrieve spatial-domain global features $Y_g^i$. The two representations, $Y_l^i$ from the spatial branch and $Y_g^i$ from the spectral branch, are subsequently fused via the Dual Fourier Attention (DFA) mechanism. This mechanism adaptively modulates cross-domain features and integrates them using learned fusion weights, as formulated in Eqs. 10–15.

This harmonious integration within the modified XceptionNet architecture enables the network to simultaneously benefit from high-resolution spatial cues and holistic spectral context, offering both computational efficiency and powerful generalization capability across diverse deepfake scenarios.

**Optimization Procedure.** Optimization of our integrated dual-domain XceptionNet architecture is performed rigorously using standard stochastic gradient descent (SGD), equipped with momentum and adaptive learning rate scheduling. To formally define our learning process, consider a labeled training set $\{(X_j, Y_j)\}_{j=1}^N$ consisting of $N$ pairs, where $X_j$ denotes an input image and $Y_j$ corresponds to its associated ground-truth label. The network's parameters $\theta$, comprising convolutional kernels, attention weights, and modulation parameters, are iteratively refined to minimize an empirical risk function $\mathcal{L}$ as follows:

$$\mathcal{L}(\theta) = \frac{1}{N} \sum_{j=1}^{N} \ell\left(Y_{\text{pred}}(X_j; \theta), Y_j\right) + \lambda R(\theta), \tag{17}$$

where $\ell$ typically denotes cross-entropy loss for classification tasks, while $R(\theta)$ represents a regularization term penalizing overly complex parameterizations and enforcing stability during training, with weight $\lambda$ balancing the empirical loss and model complexity.

At each optimization iteration $t$, the network parameters are updated via gradient descent steps, computed as follows:

$$\theta_{t+1} = \theta_t - \eta_t \nabla_\theta \mathcal{L}(\theta_t), \tag{18}$$

where $\eta_t$ is the learning rate dynamically adjusted using cosine annealing or step-wise decay to ensure convergence and mitigate local minima entrapment. Moreover, we incorporate momentum-based updates to accelerate convergence as follows:

$$v_{t+1} = \mu v_t - \eta_t \nabla_\theta \mathcal{L}(\theta_t), \quad \theta_{t+1} = \theta_t + v_{t+1}, \tag{19}$$

where a momentum factor $\mu \in [0, 1]$ controlling inertia from previous gradients.

Through meticulous backpropagation, gradients are carefully computed, taking into account intricate interactions between local convolutional layers, global spectral modules, and the DFA attention mechanism. This comprehensive optimization procedure encourages the XceptionNet architecture to learn discriminative, multi-domain representations, ultimately facilitating significant performance improvements in challenging vision tasks.

## 4 Experimental Results and Analysis

### 4.1 Implementation Details

**Datasets.** To evaluate the robustness and generalizability of our proposed framework, we conduct experiments on a diverse collection of real and synthetic datasets. For real-world relevance, we adopt widely-used datasets introduced by Wu et al. [67] and further augment our benchmarks with self-curated data synthesized via SOTA generative models. Specifically, fake samples are generated using Stable Diffusion v1.4 [48, 57], DreamBooth [45], and Latent Diffusion models from CompVis, thereby ensuring coverage of contemporary diffusion-based generation techniques.

Our evaluation encompasses synthetic data produced by eleven distinct generation pipelines, including ProGAN [33], StyleGAN2 [35], StyleGAN3 [75], BigGAN [7], EG3D [9], Taming Transformers [22],

DALL-E 2 [56], GLIDE [52], Latent Diffusion [26], Guided Diffusion [21], and Stable Diffusion v1.4. Corresponding real image distributions are sourced from large-scale and diverse datasets, namely ImageNet [20], COCO [42], and Danbooru [1].

To further assess the model's effectiveness under real-world conditions, we introduce a new practical benchmark denoted as $T_{Gen}$, comprising 1,000 images per category from DreamBooth, Midjourney v4 and v5 [27], NightCafe [59], StableAI, and YiJian [44]. Prompt engineering and generation details for these samples are presented in the Appendix. Additionally, to establish a comprehensive generalization analysis, we evaluate our model on the GenImage benchmark [74], a dataset designed to test deepfake detectors under varied synthetic domains. For evaluating diffusion-based generation, we also sample 1,000 images each from SDXL [54] and DiffusionDB [63] to ensure a robust and diverse test suite covering different model classes. Our experimental suite includes both object-centric and face-centric manipulations, spanning indoor, outdoor, and natural scenes to ensure domain diversity. The complete training corpus contains 470K real and 410K synthetic images. During testing, we uniformly sample 5,000 images per synthetic category, yielding a total of 62K fake and 62K real images in the test set. Detailed dataset splits are reported in Table 1.

**Experimental Setup.** All experiments are conducted using PyTorch (v3.6) on a system equipped with four NVIDIA Titan RTX GPUs, accelerated via CUDA 11.3. Training is performed with a batch size of 48, and all images are resized to a fixed resolution of $224 \times 224$ pixels. We use the standard cross-entropy loss to optimize classification, coupled with the Adam optimizer initialized at a learning rate of 0.1. The training process incorporates a composite learning rate schedule: an initial 5-epoch warm-up phase is followed by a cosine annealing strategy to gradually decay the learning rate. Additionally, learning rate reductions by a factor of 10 are applied at epochs 30, 60, and 80 to promote convergence. Model performance is assessed using three evaluation metrics: accuracy, Area Under the ROC Curve (AUC), and mean Average Precision (mAP) to offer a well-rounded view of detection capability.

## 4.2 Comparison with SOTA Methods

We extensively evaluated our approach on different datasets using our method which is enhanced with dual-domain processing, rather than relying on pre-trained models such as CLIP. Unlike prior methods that heavily depend on off-the-shelf feature extractors, our approach integrates spectral-spatial learning via a customized spectral cross-attentional framework. As shown in the last row in Table 2, our method achieves the highest average AUC and accuracy compared to existing SOTA approaches: Wang et al. [65], Chandrasegaran et al. [10], Chai et al. [8], Grag et al. [23], Xu Zhang et al. [71], and UGAD [4]. We find that our approach outperforms these methods in the all of cases.

In addition, recent works have explored generalization in fake image detection using vision-language models. Davide et al. [19] investigate the use of CLIP features for detecting AI-generated content and demonstrate strong generalization across models like DALL·E 3 and MidjourneyV5 with minimal training data. Similarly, Ojha et al. [51] propose a universal fake image detector that generalizes across multiple generative models, while "CLIPping the Deception" [38] adapts vision-language models for universal deepfake detection, showing their applicability across image domains. Corvi et al. [17] focus specifically on diffusion models, evaluating detection methods tailored to this emerging class of generators. These studies collectively emphasize that large domain-specific datasets are not essential, and lightweight detection pipelines can yield competitive results. We include a comparative evaluation against these approaches in Table 3, where our method consistently demonstrates superior performance in accuracy for distinguishing real and fake content except Corvi et al. in LDM method. For this evaluation, we use the GenImage dataset [74], which contains a total of 24,000 synthetic images evenly split across eight generative families. For deepfake benchmarks, we adopt the FaceForensics++ (FF++) dataset [58], utilizing 100 test videos each for Deepfakes and FaceSwap, corresponding to approximately 10,000 frames per manipulation type. Additionally, we sample 1,000 images each from diffusion-based models including Guided Diffusion, SDXL, and DiffusionDB to evaluate generalization performance across diverse diffusion datasets.

## 4.3 Ablation Studies

**Performance Scaling with Different Architecture.** Table 4 reports the incremental gains from integrating the proposed modules across ResNet [25] and XceptionNet backbones. ResNet50 sees an increase from 69.0% to 84.1%, while ResNet101 and ResNet152 improve from 78.2% and 79.9% to 90.2% and 93.0%, respectively. XceptionNet shows the strongest performance, reaching 96.4%. These results indicate that deeper networks benefit more from spectral-spatial

Table 1: Summary of datasets used for training, evaluation, and generalization testing. The "Train" and "Test" columns indicate the dataset's usage in model training and evaluation, respectively. The number of test images is set to 5,000 fewer than the training samples for entries with ✓ in both columns. All real image families include 20,000 samples in the test set. The "$T_{Gen}$" column indicates inclusion in the practical generalization benchmark.

| Family | Method | #Images | Train | Test | $T_{Gen}$ |
|---|---|---|---|---|---|
| Real | ImageNet | 140k | ✓ | ✓ | × |
| Real | MS COCO | 120k | ✓ | ✓ | × |
| Real | LSUN | 120k | ✓ | × | × |
| Real | Danbooru & Artist | 110k | ✓ | ✓ | × |
| GAN | ProGAN | 210k | ✓ | ✓ | × |
| GAN | StyleGAN2 | 5k | × | ✓ | × |
| GAN | StyleGAN3 | 5k | × | ✓ | × |
| GAN | BigGAN | 5k | × | ✓ | × |
| GAN | Eg3D | 5k | × | ✓ | × |
| Transformer | Taming Transformer | 5k | × | ✓ | × |
| Diffusion | GLIDE | 6k | × | ✓ | × |
| Diffusion | Stable Diffusion V1.4 | 210k | ✓ | ✓ | × |
| Diffusion | Latent Diffusion | 7k | × | ✓ | × |
| Diffusion | DALL-E 2 | 7k | × | ✓ | × |
| Diffusion | Guided Diffusion | 5k | × | ✓ | × |
| Diffusion | SDXL | 1k | × | ✓ | × |
| Diffusion | DiffusionDB | 1k | × | ✓ | × |
| Diffusion | DreamBooth | 1k | × | ✓ | ✓ |
| Diffusion | MidjourneyV4 | 1k | × | ✓ | ✓ |
| Diffusion | MidjourneyV5 | 1k | × | ✓ | ✓ |
| Diffusion | NightCafe | 1k | × | ✓ | ✓ |
| Diffusion | StableAI | 1k | × | ✓ | ✓ |
| Diffusion | YiJian | 1k | × | ✓ | ✓ |

Table 2: Benchmark comparison of deepfake detection methods evaluated across a comprehensive suite of GAN-based and diffusion-based generative models, including datasets derived from ImageNet, COCO, artist-rendered sources, and Danbooru. The table contrasts the performance of several SOTA baselines with our proposed SpecXNet, all trained using ProGAN and Stable Diffusion V1.4 for consistent evaluation. Here, Bold indicates best performance, while underlined denotes second-best.

| Real | Datasets Fake | | Grag AUC | mAP | Acc | CR AUC | mAP | Acc | Wang AUC | mAP | Acc | Zhang AUC | mAP | Acc | PatchFor AUC | mAP | Acc | UGAD AUC | mAP | Acc | SpecXNet (Ours) AUC | mAP | Acc |
|---|---|---|---|---|---|---|---|---|---|---|---|---|---|---|---|---|---|---|---|---|---|---|---|
| ImageNet + COCO | GAN | BigGAN | .745 | .826 | .796 | .725 | .657 | .534 | .858 | .843 | .795 | .485 | .621 | .497 | .653 | .581 | .504 | .951 | .952 | .936 | **.981** | **.965** | **.972** |
| | | StyleGAN2 | .858 | .950 | .912 | .870 | .590 | .558 | .899 | .765 | .728 | .505 | .540 | .518 | .736 | .770 | .508 | .967 | .957 | .928 | **.975** | **.966** | **.945** |
| | | StyleGAN3 | .908 | .895 | .854 | .869 | .645 | .615 | .901 | .785 | .754 | .519 | .550 | .529 | .767 | .790 | .502 | .921 | .915 | .877 | **.942** | **.930** | **.902** |
| | | ProGAN | .833 | .810 | .772 | .794 | .685 | .654 | .880 | .865 | .815 | .485 | .510 | .497 | .653 | .680 | .504 | .987 | .980 | .957 | **.993** | **.986** | **.962** |
| | | EG3D | .793 | .710 | .668 | .856 | .575 | .537 | .860 | .840 | .799 | .606 | .635 | .589 | .819 | .850 | .498 | .872 | .860 | .834 | **.974** | **.965** | **.942** |
| | DM | DALL-E 2 | .516 | .580 | .552 | .522 | .555 | .520 | .586 | .610 | .560 | .650 | .675 | .620 | .584 | .612 | .497 | .941 | .965 | .872 | **.962** | **.973** | **.945** |
| | | GLIDE | .574 | .620 | .588 | .624 | .655 | .528 | .608 | .640 | .600 | .525 | .550 | .531 | .715 | .745 | .510 | .936 | .965 | .927 | **.950** | **.970** | **.941** |
| | | Latent Diffusion | .863 | .705 | .675 | .844 | .880 | .907 | .749 | .784 | .650 | .463 | .485 | .479 | .652 | .682 | .506 | .970 | .951 | .921 | **.975** | **.980** | **.943** |
| | | Taming Transformer | .710 | .722 | .692 | .757 | .787 | .703 | .943 | .973 | .652 | .791 | .821 | .790 | .741 | .771 | .610 | .950 | .980 | .876 | **.978** | **.985** | **.940** |
| | | Stable DiffusionV1.4 | .580 | .615 | .592 | .578 | .601 | .523 | .577 | .598 | .609 | .415 | .476 | .444 | .731 | .789 | .698 | .922 | .952 | .937 | **.955** | **.970** | **.945** |
| | | Guided Diffusion | .588 | .607 | .577 | .584 | .614 | .520 | .566 | .596 | .652 | .491 | .521 | .491 | .691 | .721 | .510 | .925 | .945 | .915 | **.965** | **.965** | **.948** |
| Artist + Danbooru | GAN | BigGAN | .949 | .913 | .883 | .892 | .922 | .958 | .946 | .976 | .984 | .705 | .735 | .733 | .857 | .887 | .725 | .942 | .972 | .956 | **.986** | **.984** | **.978** |
| | | StyleGAN2 | .951 | .913 | .883 | .911 | .941 | .899 | .969 | .999 | .972 | .770 | .800 | .713 | .883 | .913 | .702 | .985 | .965 | .982 | **.989** | **.986** | **.985** |
| | | StyleGAN3 | .978 | .981 | .951 | .966 | .996 | .967 | .969 | .999 | .972 | .440 | .470 | .338 | .718 | .748 | .500 | .983 | .955 | .970 | **.990** | **.987** | **.986** |
| | | ProGAN | .970 | .974 | .899 | .935 | .980 | .986 | .952 | .980 | .991 | .992 | .997 | .835 | .976 | .988 | .649 | .992 | .997 | .992 | **.995** | **.998** | **.992** |
| | | EG3D | .823 | .847 | .722 | .826 | .850 | .787 | .905 | .925 | .943 | .803 | .828 | .896 | .500 | .525 | .784 | .962 | .984 | .985 | **.986** | **.989** | **.987** |
| | DM | DALL-E 2 | .778 | .803 | .616 | .782 | .808 | .827 | .746 | .770 | .737 | .933 | .960 | .821 | .525 | .550 | .500 | .955 | .980 | .898 | **.963** | **.985** | **.961** |
| | | GLIDE | .827 | .850 | .668 | .754 | .790 | .845 | .754 | .780 | .814 | .819 | .845 | .734 | .966 | .990 | .503 | .974 | .970 | .957 | **.981** | **.986** | **.963** |
| | | Stable DiffusionV1.4 | .765 | .800 | .659 | .767 | .807 | .726 | .901 | .931 | .882 | .742 | .885 | .701 | .832 | .865 | .659 | .962 | .965 | .936 | **.984** | **.988** | **.967** |
| | | Guided Diffusion | .869 | .885 | .670 | .818 | .850 | .897 | .716 | .735 | .681 | .793 | .810 | .769 | .831 | .850 | .644 | .988 | .995 | .974 | **.993** | **.997** | **.982** |
| | | Latent Diffusion | .863 | .695 | .675 | .844 | .875 | .908 | .749 | .770 | .650 | .785 | .805 | .721 | .843 | .865 | .678 | .971 | .980 | .959 | **.985** | **.992** | **.975** |
| | | Taming Transformer | .865 | .670 | .651 | .878 | .900 | .931 | .966 | .985 | .915 | .866 | .890 | .796 | .694 | .715 | .545 | .969 | .945 | .960 | **.987** | **.993** | **.982** |
| | | Average | .799 | .770 | .729 | .787 | .815 | .746 | .815 | .850 | .781 | .662 | .690 | .637 | .738 | .770 | .570 | .957 | .963 | .935 | **.985** | **.989** | **.978** |

Table 3: Comprehensive comparison of deepfake detection performance across diverse generative models. Evaluation is conducted on GenImage (covering both GAN-based and diffusion-based generators), recent diffusion techniques, and traditional face manipulation datasets (FF++). The table contrasts the performance of several SOTA baselines with our proposed SpecXNet. All methods, including our SpecXNet, are trained using ProGAN and Stable Diffusion V1.4 to ensure fairness in evaluation. Accuracy is reported across each category, with the final column representing the overall average across all benchmarks.

| Method | GenImage | | | | | | | | Diffusion Methods | | | | | FF++ | | Average |
|---|---|---|---|---|---|---|---|---|---|---|---|---|---|---|---|---|
| | BigGAN | SD V1.4 | SD V1.5 | ADM | GLIDE | Wukong | VQDM | Midjourney | Glide | Guided | LD | SDXL | DiffusionDB | DeepFakes | FaceSwap | |
| Davide et al. | .868 | .851 | .806 | .824 | .843 | .837 | .819 | .826 | .564 | .551 | .611 | .629 | .580 | .527 | .497 | .706 |
| Corvi et al. | .883 | .897 | .874 | .863 | .859 | .832 | .821 | .847 | .598 | .509 | **.973** | .891 | .901 | .593 | .501 | .786 |
| Ojha et al. | .902 | .910 | .896 | .851 | .889 | .871 | .865 | .883 | .754 | .895 | .892 | .822 | .848 | .799 | .602 | .825 |
| Khan et al. | .928 | .930 | .917 | .905 | .899 | .882 | .898 | .906 | .915 | .929 | .847 | .778 | .751 | .784 | .747 | .868 |
| Ours | **.965** | **.978** | **.963** | **.932** | **.958** | **.949** | **.946** | **.958** | **.960** | **.937** | .947 | **.902** | **.904** | **.832** | **.824** | **.930** |

modeling and attention-based fusion. The consistent improvements across architectures confirm that the proposed modules generalize well, significantly enhancing representational power.

**Effectiveness of Our Proposed Components.** The proposed DDFC contributes significantly to the model's ability to discern subtle yet global inconsistencies present in manipulated imagery. Unlike conventional convolutional backbones that predominantly capture spatial features, DDFC explicitly decomposes the input into local and spectral pathways, enabling the capture of periodic artifacts and frequency distortions common in synthetic content. Empirical results in Table 4 reveal that incorporating DDFC alone leads to consistent improvements across all backbone networks, with average gains exceeding 10% in many cases. These findings indicate that spectral cues when modeled through structured decomposition, provide a powerful inductive bias for forgery detection.

The DFA mechanism serves as a bridge between the spatial and spectral domains by dynamically modulating their interactions. Rather than naively concatenating features from distinct domains, DFA introduces a cross-attention mechanism that selectively amplifies or suppresses local and global features based on content-aware importance. This results in more discriminative representations and enhanced robustness to diverse manipulations. When applied independently, DFA builds upon the DDFC's output by refining the fusion process, yielding measurable performance boosts, especially in deeper architectures where semantic granularity is richer.

When jointly applied, DDFC and DFA form a synergistic framework that capitalizes on their respective strengths: DDFC's ability to encode orthogonal domain features and DFA's capacity for adaptive feature alignment. The dual-branch formulation of DDFC ensures comprehensive representation coverage, while DFA optimally blends these features into a unified embedding space. As reflected in the substantial accuracy gains across all evaluated architectures, including a peak performance of 96.4% on XceptionNet, this combination yields a model that is not only highly performant but also generalizes well to unseen manipulations. The architecture demonstrates the efficacy of dual-domain modeling paired with dynamic attention in advancing the state of deepfake detection.

Table 4: Impact of ResNet architecture and SpecXNet components (DDFC and DFA) across different generation methods. SG2 denotes StyleGAN2, LD refers to Latent Diffusion. 'Average' indicates mean performance across all test datasets.

| Backbone | Configuration | Accuracy on Generation Methods | | | | Average |
|---|---|---|---|---|---|---|
| | | ImageNet + SG2 | ImageNet + DALL-E 2 | Artist+Danbooru + GLIDE | Artist+Danbooru + LD | |
| ResNet50 | None | .713 | .679 | .697 | .681 | .690 |
| ResNet50 | DDFC | .837 | .792 | .814 | .814 | .817 |
| ResNet50 | DDFC + DFA | .851 | .807 | .836 | .854 | .841 |
| ResNet101 | None | .796 | .751 | .792 | .764 | .782 |
| ResNet101 | DDFC | .895 | .842 | .901 | .909 | .887 |
| ResNet101 | DDFC + DFA | .901 | .848 | .929 | .912 | .902 |
| ResNet152 | None | .832 | .778 | .798 | .788 | .799 |
| ResNet152 | DDFC | .905 | .857 | .912 | .918 | .928 |
| XceptionNet | None | .847 | .794 | .816 | .802 | .813 |
| XceptionNet | DDFC | .912 | .888 | .921 | .927 | .936 |
| **XceptionNet** | **DDFC + DFA** | **.945** | **.945** | **.963** | **.975** | **.964** |

Table 5: Evaluation of cross-domain generalization on the practical benchmark dataset $T_{Gen}$, measured by classification accuracy. Each model is trained using our dataset and evaluated on six distinct generative subsets. Wang-0.1 and Wang-0.5 denote models trained using 10% and 50% data augmentation, respectively.

| Method | $T_{gen}$-Accuracy | | | | | | |
|---|---|---|---|---|---|---|---|
| | Dream Booth | Midjourney V4 | Midjourney V5 | Night Cafe | Stable AI | Yi Jian | Average |
| Wang-0.5 | .839 | .829 | .795 | .811 | .800 | .729 | .801 |
| Wang-0.1 | .847 | .861 | .881 | .867 | .832 | .715 | .833 |
| CR | .764 | .751 | .725 | .790 | .770 | .582 | .731 |
| Grag | .655 | .682 | .665 | .731 | .729 | .537 | .667 |
| **Ours** | **.966** | **.929** | **.893** | **.952** | **.948** | **.821** | **.900** |

**Generalization Capability on Practical Dataset $T_{Gen}$.** To assess the robustness of our model under real-world distribution shifts, we evaluate generalization performance on the $T_{Gen}$ dataset, which comprises diverse samples generated from DreamBooth, Midjourney V4/V5, NightCafe, StableAI, and YiJian. As shown in Table 5, our model consistently outperforms prior SOTA methods across all six subdomains, achieving an average accuracy of 90.0%. Compared to the strongest baseline (Wang-0.5), which records 80.1%, our approach yields a substantial gain of 9.9%. Notably, our model surpasses others even on challenging generators such as YiJian and Midjourney V5, reflecting strong resilience to stylistic and semantic variation. This superior performance highlights the effectiveness of our dual-domain representation and adaptive attention mechanism in capturing generalizable forgery patterns.

**Robustness Against Post-Processing Artifacts.** To assess our detector's resilience under real-world post-processing, we evaluate its generalization on the practical dataset $T_{Gen}$ across eleven common augmentations (Figure 3), including *Blur*, *JPEG compression*, their combinations *Blur+JPEG (0.5)* and *Blur+JPEG (0.1)*, and other transformations like *ResizedCrop*, *ColorJitter*, *Rotation*, *Affine*,

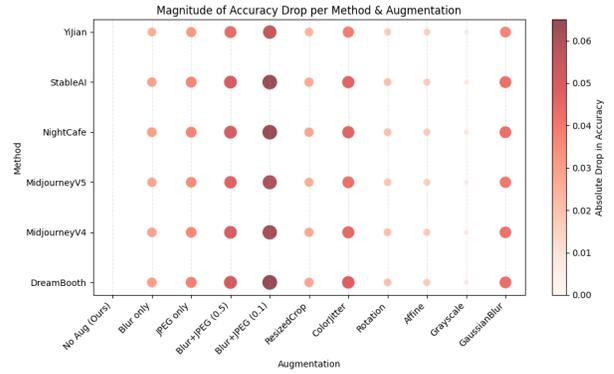

Figure 3: Effect of augmentations on detector performance across the $T_{Gen}$ dataset. Models trained on ProGAN are evaluated on unseen generators. While most augmentations enhance robustness, certain models like MidjourneyV5 show reduced accuracy under specific perturbations.

*Grayscale*, and *GaussianBlur*, compared to a *No Augmentation (Ours)* baseline. Our model retains high accuracy across these variations, though dual-frequency corruptions like Blur+JPEG cause notable drops—e.g., DreamBooth accuracy decreases from 0.966 to 0.915 and 0.901 at 0.5 and 0.1 probabilities. MidjourneyV4 and V5 also suffer over 6% degradation, reflecting their sensitivity to low-frequency noise. In contrast, geometric and photometric changes (e.g., Rotation, ColorJitter) result in smaller drops (under 2–3%). YiJian remains the hardest generator, dropping from 0.821 to 0.766 under Blur+JPEG (0.1), indicating subtler artifacts. Nevertheless, our dual-domain model maintains over 90% accuracy in most settings. This robustness arises from the architecture and Dual Fourier Attention (DFA): the spectral branch captures frequency distortions, the spatial branch handles geometric noise, and the DFA adaptively fuses both. Thus, our method reliably detects forgeries across generators and post-processing variations, supporting deployment in unconstrained scenarios.

Overall, the synergy of DDFC and DFA modules enhances architectural scalability and drives generalization across varied deepfake generation methods with different augmentation techniques.

## 5 Conclusion

In this work, we introduced the SpecXNet, a novel dual-domain deepfake detection framework that integrates spatial and spectral cues through a unified convolutional architecture. By explicitly decomposing feature representations into local spatial and global spectral branches, and fusing them via the proposed DFA mechanism, DDFC effectively captures both fine-grained textural anomalies and long-range frequency inconsistencies inherent in synthetic media. Our extensive evaluations across diverse benchmarks, including GAN, diffusion, and real-world generative models, demonstrate that our proposed SpecXNet achieves SOTA performance in accuracy and generalization, even under challenging post-processing perturbations. These results highlight the critical importance of joint spatial-spectral modeling and content-aware attention for robust and scalable deepfake detection.

## Acknowledgments

This work was partly supported by Institute for Information & communication Technology Planning & evaluation (IITP) grants funded by the Korean government MSIT: (RS-2022-II221199, RS-2022-II220688, RS-2019-II190421, RS-2023-00230337, RS-2024-00356293, RS-2024-00437849, RS-2021-II212068, RS-2025-02304983, and RS-2025-02263841).

# (Supplementary Materials)
# SpecXNet: A Dual-Domain Convolutional Network for Robust Deepfake Detection

Inzamamul Alam
Department of Computer Science and Engineering
Sungkyunkwan University
Suwon, Republic of Korea
inzi15@g.skku.edu

Md Tanvir Islam
Department of Computer Science and Engineering
Sungkyunkwan University
Suwon, Republic of Korea
tanvirnwu@g.skku.edu

Simon S. Woo*
Department of Computer Science and Engineering
Sungkyunkwan University
Suwon, Republic of Korea
swoo@g.skku.edu

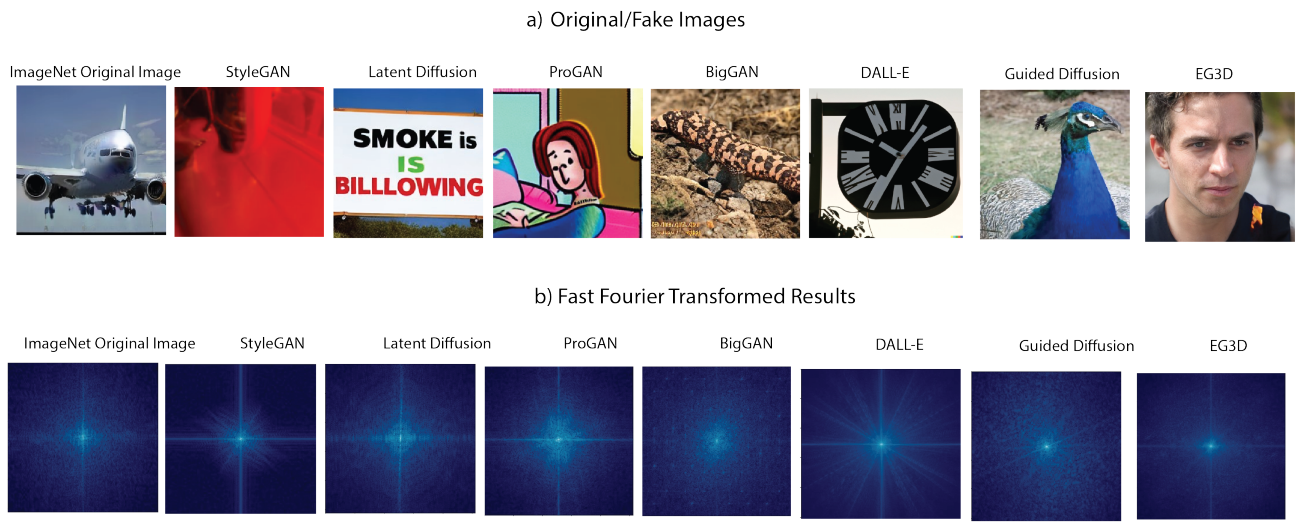

Figure 1: Examples of the AI-generated fake images and their corresponding FFT results: (Top) images generated by state-of-the-art generative models: StyleGAN3 [10], Latent Diffusion [4], ProGAN [6], BigGAN [1], DALL-E [8], Guided Diffusion [3], and EG3D [2], respectively. The bottom images have corresponding FFT results that have specific frequencies corresponding to each generated model.

## 1 FFT Signatures of Generative Models

Figure 1 showcases visual comparisons between original ImageNet images and outputs from various state-of-the-art generative models, including StyleGAN, Latent Diffusion, ProGAN, BigGAN, DALL-E, Guided Diffusion, and EG3D. The top row presents synthesized samples from each generator, while the bottom row illustrates the corresponding 2D Fast Fourier Transforms (FFT) of these images.

Notably, each generative model produces a distinct frequency signature in the Fourier domain. While natural images exhibit dense but smoothly decaying high-frequency spectra, AI-generated samples reveal unique structural patterns—such as radial lines, symmetry artifacts, and spectral voids—arising from upsampling heuristics, architectural inductive biases, or latent priors.

These frequency-domain discrepancies provide a powerful modality for manipulation detection, motivating our dual-domain architecture. The Spectral Fusion framework explicitly models such artifacts by decomposing input features into spatial and spectral branches, allowing our model to attend to both texture-level anomalies and frequency-based inconsistencies. As highlighted in this figure, leveraging FFT-based representations reveals generator-specific spectral traces that are often imperceptible in the spatial domain.

## 2 Additional Information on Dataset

We have used many prompts to generate images using stable diffusion V1.4 [9], midjourney V5 [5], Dreambooth [7], etc. Here are some samples of the prompts that we used for image generation.

(1) Bruce Lee sitting in a car on a road way.
(2) Mother Teresa portrait from front, she is serious with a beautiful makeup.
(3) Cyberpunk sci-fi woman portrait sitting in a pipe light session.
(4) An angry 6 year old girl steering in jungle cloths having burn marks on her face.

---
*Corresponding author. Email: swoo@g.skku.edu (Simon S. Woo)

Table 1: Complexity analysis of SpecXNet. We report parameter and FLOPs estimates for the vanilla Xception block, the proposed SpecXNet components—including the Dual-Domain Feature Coupler (DDFC), Spectral Transform, and Dual Fourier Attention (DFA)—as well as the total per-block cost. $C_1, C_2$ are input/output channels, $H \times W$ is spatial resolution, $K$ is kernel size, and $\alpha$ determines the global channel ratio.

| Module | Parameter Count | FLOPs Estimate |
|---|---|---|
| **Baseline (XceptionNet)** | $C_1 C_2 K^2$ | $C_1 C_2 K^2 HW$ |
| **DDFC** | | |
| Local → Local | $(1-\alpha)^2 C_1 C_2 K^2$ | $(1-\alpha)^2 C_1 C_2 K^2 HW$ |
| Global → Global (Spectral) | $\alpha^2 C_2 \left(\frac{1}{2} C_1 + \frac{3}{2} C_2\right)$ | $\alpha^2 C_2 HW \left(\frac{1}{2} C_1 + \frac{13}{16} C_2\right)$ |
| Local → Global | $\alpha(1-\alpha) C_1 C_2 K^2$ | $\alpha(1-\alpha) C_1 C_2 K^2 HW$ |
| Global → Local | $\alpha(1-\alpha) C_1 C_2 K^2$ | $\alpha(1-\alpha) C_1 C_2 K^2 HW$ |
| **DFA** | $2(C_1 + C_2)^2$ | $2(C_1 + C_2)^2 + (C_1 + C_2)HW$ |
| **Total (SpecXNet Block)** | $(1-\alpha^2) C_1 C_2 K^2 + \alpha^2 C_2 \left(\frac{1}{2} C_1 + \frac{3}{2} C_2\right) + 2(C_1 + C_2)^2$ | $(1-\alpha^2) C_1 C_2 K^2 HW + \alpha^2 C_2 HW \left(\frac{1}{2} C_1 + \frac{13}{16} C_2\right) + 2(C_1 + C_2)^2 + (C_1 + C_2)HW$ |

Table 2: Efficiency comparison of SpecXNet and ResNet152+UGAD on RTX 3090 and GTX 1660 Ti. SpecXNet achieves significantly higher throughput, lower latency, and reduced VRAM usage across both platforms.

| Method | IPS ↑ (RTX 3090) | IT↓ (RTX 3090) | VRAM (GB) (RTX 3090) | IPS ↑ (GTX 1660 Ti) | IT↓ (GTX 1660 Ti) | VRAM (GB) (GTX 1660 Ti) |
|---|---|---|---|---|---|---|
| ResNet152 + UGAD | 46 | 21.6 | 2.5 | 12 | 84.3 | 2.3 |
| **SpecXNet (ours)** | **112** | **6.3** | **1.2** | **33** | **23.3** | **1.1** |

(5) Visualize a mountain top villa with trees growing on top of it. A man is standing in the building to watch the beautiful landscape.
(6) An old kitten portrait in a woman cold weather dress.
(7) Generate an image of a bustling space station with diverse inhabitants and advanced technology, inspired by the text.
(8) Realistic, hyper resolution, joker turned into vampire, laughing portrait.
(9) A canal passing through jungle covered with trees.
(10) A dog Pirate cartoon standing with his cap sailing through the sea.
(11) Pikachu standing on a rock on sea side.
(12) A Sikh bodybuilder handsome beard man portrait.
(13) Anime-style Japanese woman eating super in
(14) A front portrait of an Anime-style Japanese boy crying in his Japanese room, his face is read from his nose and cheeks.
(15) A street cat walking ahead on the cyberpunk street.

## 3 Computational Efficiency

Although our proposed methodology introduces an intricate dual-domain representation framework and adaptive attention mechanisms, its computational complexity remains highly manageable, largely due to strategic algorithmic choices and efficient spectral-domain operations. Specifically, the global spectral branch capitalizes extensively on the computational efficiency of the Fast Fourier Transform (FFT), a cornerstone algorithm whose complexity is well-studied and thoroughly optimized in numerical libraries.

Formally, the computational complexity for a two-dimensional FFT on an input feature map of dimensions $H \times W$ with $\alpha C$ spectral channels can be analytically expressed as follows:

$$O(\alpha C \cdot H \cdot W \cdot \log(HW)), \quad (1)$$

where $\alpha$ denotes the proportion of channels allocated specifically to spectral domain operations. Importantly, the $\log(HW)$ scaling factor underscores a significant efficiency advantage relative to traditional spatial convolutions that typically exhibit quadratic computational complexity relative to the kernel size. Thus, even with increasing spatial resolution or deeper layers that traditionally impose computational strain, the FFT-based spectral processing retains high efficiency.

To precisely quantify and underscore these efficiency gains, consider a traditional convolutional operation characterized by kernel size $k \times k$ across $\alpha C$ input channels producing similar global receptive fields. The conventional convolution complexity scales quadratically as:

$$O\left(\alpha C \cdot k^2 \cdot H \cdot W\right), \quad (2)$$

which quickly becomes computationally prohibitive as receptive fields and kernel sizes increase to capture large-scale global contexts. In stark contrast, the spectral approach, through its frequency-domain computation, effectively sidesteps this quadratic scaling by transforming spatial convolutions into efficient element-wise multiplications in frequency space. The complexity reduction achieved by this spectral method is profound, especially for large spatial dimensions or extensive receptive fields, typically required to capture global information.

Further enhancing efficiency, the spectral modulation step—where frequency coefficients are adaptively weighted—employs lightweight operations. Specifically, spectral modulation involves learned element-wise multiplicative operations on frequency components, amounting to a complexity of:

$$O(\alpha C \cdot H \cdot W), \quad (3)$$

significantly lower than spatial convolutions. Additionally, the inverse FFT that maps modulated global spectral features back to the spatial domain incurs computational complexity similar to the forward FFT, preserving the overall complexity advantage:

$$O\left(\alpha C \cdot H \cdot W \cdot \log(HW)\right). \tag{4}$$

Moreover, our Dual Fourier Attention (DFA) mechanism introduces minimal computational overhead by employing efficient global pooling and compact linear transformations. Global average pooling (GAP), used to summarize spatial feature maps, scales linearly with spatial resolution:

$$O\left(C \cdot H \cdot W\right), \tag{5}$$

and subsequent linear transformations for attention calculation scale merely linearly with channel dimensionality:

$$O(C^2), \tag{6}$$

which is negligible relative to convolutional layers in deep architectures.

These theoretical estimates are further validated by the empirical accounting provided in Table 1, which details the parameter and FLOP costs for each dual-domain component in SpecXNet. The local-to-local and cross-branch convolutional paths scale quadratically with the kernel size, as in traditional CNNs. In contrast, the spectral transform path replaces large-kernel spatial convolutions with FFT/iFFT operations and frequency-domain modulation, achieving favorable $O(\log HW)$ scaling relative to spatial size.

Importantly, the table captures practical design elements such as the dual-branch decomposition in the DDFC, the cost of spectral modulation, and the full complexity of the Dual Fourier Attention (DFA) module. Despite the added modeling capacity, the DFA introduces only a minor overhead ($O((C_1 + C_2)^2)$), as it relies on global average pooling and lightweight linear projections. When aggregated, the full SpecXNet block maintains computational efficiency competitive with the vanilla XceptionNet baseline while offering significantly enhanced representational power.

Furthermore, we conducted experiments for a runtime comparison between SpecXNet and ResNet152+UGAD on both RTX 3090 and GTX 1660 Ti, as presented in Table 2. SpecXNet achieves higher throughput, lower inference time, and reduced memory usage on both platforms, demonstrating its efficiency and suitability for real-time applications.

Table 3: Accuracy across different values of $\alpha$, controlling the spectral-to-spatial channel ratio in DDFC.

| $\alpha$ | 0.0 (all spatial) | 0.25 | 0.50 | 0.75 |
|---|---|---|---|---|
| **Acc (%)** | 87.9 | 92.7 | **96.3** | 93.2 |

## 4 Ablation Study on $\alpha$ Ratio in DDFC

We performed an ablation study on the spectral-to-spatial channel ratio $\alpha$ in the Dual-Domain Feature Coupler (DDFC). As shown in Table 3, setting $\alpha = 0.5$ provided the highest accuracy, suggesting that a balanced division between the spatial and spectral pathways offers the most effective feature representation. Fully spatial or heavily unbalanced configurations result in lower performance.

## 5 Code Availability

To ensure reproducibility, we released the full code on GitHub.